\documentclass{article}

\usepackage{microtype}
\usepackage{graphicx}
\usepackage{subfigure}
\usepackage{booktabs} % for professional tables
\usepackage{svg}
\usepackage{tcolorbox}

\usepackage{hyperref}

\usepackage[accepted]{icml2024}
\usepackage{amsmath}
\usepackage{amssymb}
\usepackage{mathtools}
\usepackage{amsthm}

\usepackage[capitalize,noabbrev]{cleveref}

\theoremstyle{plain}

\theoremstyle{definition}

\theoremstyle{remark}

\usepackage[textsize=tiny]{todonotes}

\icmltitlerunning{NuNER: Entity Recognition Encoder Pre-training via LLM-Annotated Data}

\begin{document}

\twocolumn[
\icmltitle{NuNER: Entity Recognition Encoder Pre-training \\ via LLM-Annotated Data}



\icmlsetsymbol{equal}{*}

\begin{icmlauthorlist}
\icmlauthor{Sergei Bogdanov}{NuMind}
\icmlauthor{Alexandre Constantin}{NuMind}
\icmlauthor{Timothée Bernard}{LLF}
\icmlauthor{Benoit Crabbé}{LLF}
\icmlauthor{Etienne Bernard}{NuMind}
\end{icmlauthorlist}

\icmlaffiliation{NuMind}{NuMind}
\icmlaffiliation{LLF}{Universit\'e Paris Diderot}

\icmlcorrespondingauthor{Etienne Bernard}{etienne@numind.ai}
\icmlcorrespondingauthor{Benoit Crabbe}{benoit.crabbe@u-paris.fr}

\icmlkeywords{Machine Learning, Natural Language Processing, Named Entity Recognition, Few-Shot learning, LLM, Artificial Data Generation, ICML}

%FOR ARXIV - WILL NEED TO BE MODIFIED FOR ICML%

\begin{center}
\textsuperscript{1} NuMind \textsuperscript{2} Universit\'e Paris Diderot

\begin{footnotesize}
    \texttt{sergei@numind.ai, alexandre@numind.ai, timothee.bernard@u-paris.fr, benoit.crabbe@u-paris.fr, etienne@numind.ai}
\end{footnotesize}
\end{center}
\vskip 0.25in %0.3in need to put back for ICML
]




% Format: 8 pages as main paper, with unlimited pages for references and appendix. (https://icml.cc/Conferences/2024/CallForPapers)

\newcommand{\NuNER}{NuNER} % So that we can easily change all occurences. For example as "{\tt NuNER}".

\begin{abstract}
Large Language Models (LLMs) have shown impressive abilities in data annotation, opening the way for new approaches to solve classic NLP problems. In this paper, we show how to use LLMs to create \NuNER{}, a compact language representation model specialized in the Named Entity Recognition (NER) task. \NuNER{} can be fine-tuned to solve downstream NER problems in a data-efficient way, outperforming similar-sized foundation models in the few-shot regime and competing with much larger LLMs. We find that the size and entity-type diversity of the pre-training dataset are key to achieving good performance. We view \NuNER{} as a member of the broader family of \textit{task-specific foundation models}, recently unlocked by LLMs.
\end{abstract}

\section{Introduction}

Named Entity Recognition (NER) --- the generic task of extracting and classifying entities from text --- is a core component of natural language processing, present in a variety of applications such as medical coding, financial news analysis, or legal documents parsing \citep{francis2019transfer, dozier2010named}. Such application typically involves solving a particular NER problem, for a particular set of entity types, thus requiring the creation of a custom model.

For the last five years, the standard procedure for creating such custom model has consisted of using a transformer encoder \citep{transformer} pre-trained in a self-supervised way to satisfy a masked language modeling (MLM) objective, such as models of the BERT family \citep{devlin2019bert, roberta, deberta}. This \emph{foundation model} is then fine-tuned in a supervised way on human-annotated data, either using a simple token-classification approach, or a more advanced strategy \citep{tadner, binder}.

% Thanks to their MLM objective and their large and diverse pre-training dataset, BERT-like models provide contextualized word representations that are well suited for NER; they obtain better results for this task than when training a model from scratch or when using context-free word representations instead [cite BERT, word2vec, glove]. Nevertheless, fine-tuning these foundation models to a target problem still requires a substantial amount of annotation in practice, especially to reach a high level of performance [cite]. Our work aims at solving this issue.

\begin{figure}[t]
\vskip 0.in
\begin{center}
\centerline{\includegraphics[width=\columnwidth]{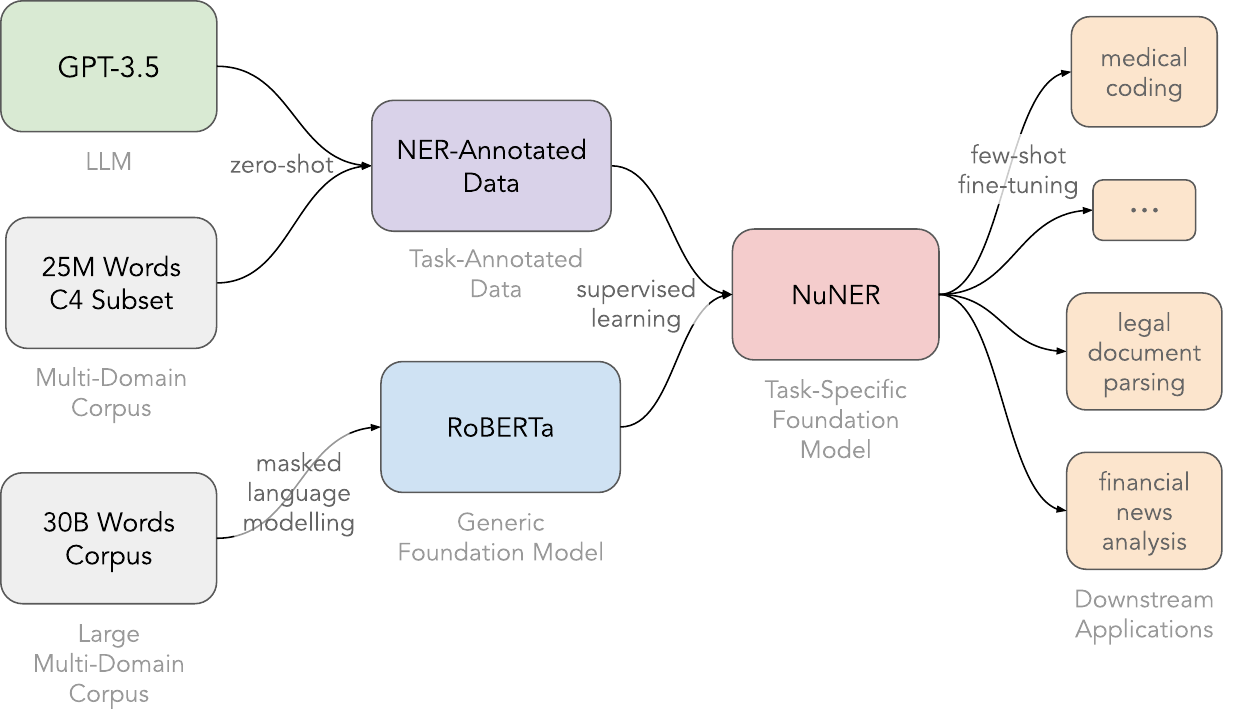}}
\vskip -0.1in
\caption{\NuNER{} creation procedure. RoBERTa is further pre-trained on a subset of C4 automatically annotated by GPT-3.5. The resulting model can be fine-tuned on various downstream NER problems.}
\label{procedure}
\end{center}
\vskip -0.1in
\end{figure}

In the last few years, we have witnessed the emergence of generative large language models (LLMs) such as GPT-3 \citep{gpt3} and, more recently, GPT-4 \citep{openai2023gpt4}, which typically have between 100 times and 10,000 times more parameters than BERT. These massive auto-regressive transformer models, trained via a next-word prediction objective, are language generators which can be prompted to perform a variety of tasks. For example, these models can be directly used through a well crafted prompt to tackle a particular NER problem with satisfying performance \citep{wang2023gptner}. The main issue with this approach is the high inference cost due to the size of LLMs.

A simple solution to this inference-cost issue is to use a correctly-prompted LLM to annotate data for the particular NER problem, and then to train a smaller model on this data. LLMs have been shown to outperform crowdworkers \citep{annot} on some tasks for a fraction of the cost so this strategy is sound, but it has issues as well. First, crafting a good prompt --- like delegating a task to someone else --- is not easy; it requires multiple back-and-forth while validating the performance using human-annotated data. Second, LLMs are not perfect annotators either \citep{mao2023gpteval}. Finally, the best LLMs are mainly hosted on private companies' servers and accessible through external APIs, opening the door to potential confidentiality and privacy leaks.

We propose an alternative approach that leverages LLMs to reduce the amount of human annotations needed to create custom models. Instead of using an LLM to directly annotate a particular single-domain dataset for a particular NER problem, our idea is to use this LLM to annotate a multi-domain dataset for variety of NER problems. We then further pre-train a small foundation model, such as BERT, on this annotated dataset. The resulting pre-trained model can then be fine-tuned to any downstream NER problem, just like any other foundation model, as depicted in \cref{procedure}.

Because the resulting pre-trained model is specialized to a generic task but is still meant to be fine-tuned to a particular problem, we refer to such a model as a \textit{task-specific foundation model}. Note that compact domain-specific foundation models like SciBERT \citep{scibert} or BioBERT \citep{biobert} are common, but task-specific foundation models of this kind are rare, mostly due to the lack of suitable datasets. Generative LLMs are the key to building such models.

In this paper, we apply the above idea to create \NuNER{}, a task-specific foundation model for the generic task of NER.

In \cref{NuNER_Sec}, we describe both the dataset creation and the training procedures. In a nutshell, we use GPT-3.5 to annotate a subset of C4 \citep{c4}, resulting in a 24.4M words dataset containing 4.38M annotations from 200k unique concepts. We then pre-train a base RoBERTa on this dataset via a contrastive-learning approach \citep{contrastive_learning} to obtain \NuNER{}.

In \cref{TLPerf}, we analyze the transfer learning performance of \NuNER{} in an extended few-shot regime. We find that \NuNER{} largely outperforms both its base model and the same base model further pre-trained on NER-BERT data \citep{ner-bert}, which is the largest and most diverse NER dataset we could find. These results demonstrate the validity of our approach.

In \cref{study}, we investigate the factors influencing \NuNER{}'s abilities. We find that the diversity of the annotations and the size of the pre-training dataset are the most influential factors. Surprisingly, the diversity of the text does not appear to be as influential.

In \cref{LLMSec}, for informational purposes, we compare fine-tuning \NuNER{} with using GPT-3.5 and GPT-4 via in-context learning. We find that \NuNER{} beats GPT-3.5 and competes with GPT-4 when more than a dozen entities of each type is seen during training. We also compare, via fine-tuning, \NuNER{} with UniversalNER \citep{zhou2023universalner}, a recent LLM specialized in the NER task. We find that they exhibit similar transfer learning performance when fine-tuned, despite \NuNER{} being 56 times smaller.

The contributions of our paper are as follows.

\textbf{1.} We introduce and demonstrate the validity of a procedure that consists of annotating raw data with an LLM in order to train a task-specific foundation model for NER.

\textbf{2.} We identify the factors that are likely to improve the performance of the resulting task-specific foundation model.

\textbf{3.} We provide and open-source \NuNER{}\footnote{\label{link} https://huggingface.co/numind %https://www.will\_be\_inserted\_after\_acceptance.com.
}, a compact encoder-based language representation model for NER. \NuNER{} outperforms similar-sized models, competes with LLMs, and can be used as a drop-in replacement for RoBERTa.

\textbf{4.} We provide and open-source an LLM-annotated NER dataset\footref{link}, containing 4.38M annotations from 200k entity types, which is suitable for pre-training NER models.

\section{Related Work}
\label{related}

Early attempt to create NER-specific foundation models for low-resource NER problems focused on leveraging Wikipedia anchors \citep{mengge-etal-2020-coarse, cao-etal-2019-low}. In particular, \citet{ner-bert} (NER-BERT) combined these anchors and DBpedia Ontology to create a NER dataset containing 3.64M entities from 315 entity types. BERT was then pre-trained on this dataset, leading to improved few-shot performance. We demonstrate in \cref{FSFrozen} that our LLM-annotation procedure outperforms such approach.

Subsequent work has focused on pre-training generative LLMs on a large number of existing human-annotated NER datasets to achieve strong zero-shot capabilities \citep{wang2023instructuie, sainz2023gollie}. These works differ from ours in terms of type of data used, model employed, and objective (zero-shot vs. few-shot).

Recently, \citet{zhou2023universalner} proposed UniversalNER, an LLM with 7B and 13B parameters, also pre-trained on data annotated by GPT-3.5. Our work primarily differ in the model and training procedure: we found a way to train an encoder model with only 125M parameters on such data, making it substantially cheaper to use. In \cref{LLMSec}, we show that \NuNER{} and UniversalNER have similar transfer-learning abilities when provided with more than a few training examples per entity types.

Even more recently, \citet{zaratiana2023gliner} proposed GLiNER, which uses UniversalNER's data to pre-train a small encoder model. Our work was done concurrently and independently from this work. The principal distinction is in the architecture used: GLiNER merges the text and concept encoders, whereas in our approach, they remain independent (see \cref{model-schema}). While GLiNER's integrated approach likely enhances performance, our decision to keep the encoders separate enables \NuNER{} to function as a concept-agnostic language representation model. This gives the possibility to pre-compute text embeddings for a variety of applications such as information retrieval. This also makes \NuNER{} a viable drop-in substitute for BERT or RoBERTa in standard NER methodologies.

\section{\NuNER{}}
\label{NuNER_Sec}

The creation of \NuNER{} is a two-step process: dataset creation and model training.

\subsection{Dataset Creation}
\label{data_creation}

We begin with a random sample of C4 \citep{c4}, an English web crawl corpus that contains text from a wide range of sources, including blog posts, news articles, and social media messages. We selected this dataset for its domain diversity.

We want to annotate this dataset with entities spanning a large and diverse set of types in order for our model to generalize to all kind of NER problems. To achieve this we opt for an unconstrained approach: we allow the LLM to extract any entity it identifies, and give it the freedom to assign any type it deems appropriate for each entity. This includes annotating with concepts more akin to topics than entity types (e.g., ``wellness"). We refer to these entity types/topics as \emph{concepts}. To resolve potential ambiguities, we also ask the LLM to provide descriptions for the concepts it identifies. However, we ultimately disregarded these descriptions as we found that they do not improve \NuNER{}'s performance. Our prompt is shown in \cref{prompt}.

\begin{figure}[h]
    \vskip 0.in
    \begin{small}
    \begin{tcolorbox}[width=\columnwidth, colback=green!2, boxrule=0.01in]
The goal is to create a dataset for entity recognition. Label as many entities, concepts, and ideas as possible in the input text. Invent new entity types that may not exist in traditional NER Tasks such as more abstract concepts and ideas. Make sure the entity concept is not part of speech but something more meaningful. Avoid finding meaningless entities. \newline
Output format (separate entities with new line):
\newline
entity from the text $<>$ entity concept $<>$ description of entity group/concept \newline
Input: \newline
\begin{footnotesize} \sc{[input sentence]} \end{footnotesize}
\end{tcolorbox}
\vskip -0.1in
 \end{small}
\caption{Prompt used to annotate \NuNER{}'s pre-training data.}
\label{prompt}
\vskip -0.1in
\end{figure}

Note that we do not ask to return the position of the entity in the text as LLMs are not good at counting. To train \NuNER{}, we have to retrieve this position through an exact string match, which can lead to annotation errors in rare cases.

We use gpt-3.5-turbo-0301 with this prompt to annotate 1.35M sentences. \cref{example} shows one of these annotated sentences. We then apply a simple filter to remove sentences containing an annotation with the concept ``concept", as we consider this too uninformative, and obtain a final dataset of 1M annotated sentences.

\begin{figure}[h]
    \vskip 0.in
     \begin{tcolorbox}[width=\columnwidth, colback=blue!2, boxrule=0.01in]
    \begin{small}
    ``Steven Means has signed a one-year contract extension with the Falcons after making four starts in 2018."
    \begin{center}
    \begin{tabular} {ccccc} % {lcccr}
    \sc{entity} & \sc{concept} & \sc{description} \\
    \midrule
    Steven Means & NFL player & professional athlete \\
    \midrule
    Falcons & NFL team & professional sports team \\
    \midrule
    2018 & year & unit of time \\
    \end{tabular}
    \end{center}
    \end{small}
\end{tcolorbox}
    \vskip -0.1in
    \caption{Sentence from C4 annotated with GPT-3.5.}
    \label{example}
    \vskip -0.1in
\end{figure}

We find that GPT-3.5 performs quite well in this task. From a manual review of 100 examples, we estimate that, in more than 95\% of cases, the concept associated with the extracted entity is completely sensible, as shown in \cref{example}. In fewer that 5\% of cases, the extraction and associated concept are questionable, such as ``cosmos seeds" identified as ``plant variety" when it is actually a seed. However, many entities are missed. For instance, in the example of \cref{example}, the LLM could have also identified ``person". In a sense, this method has high precision but low recall. In \cref{Model_traninig}, we propose a training procedure that mitigates the impact of these false negatives.

The resulting dataset comprises a total of 4.38M entity annotations, distributed across 200k unique concepts. These concepts cover a wide range of domains, as illustrated in \cref{concept-map}. The diversity of concepts in this dataset is far greater than what can be found in human-annotated NER datasets.

\begin{figure}[h]
\vskip 0.0in
\begin{center}
\centerline{\includegraphics[width=\columnwidth]{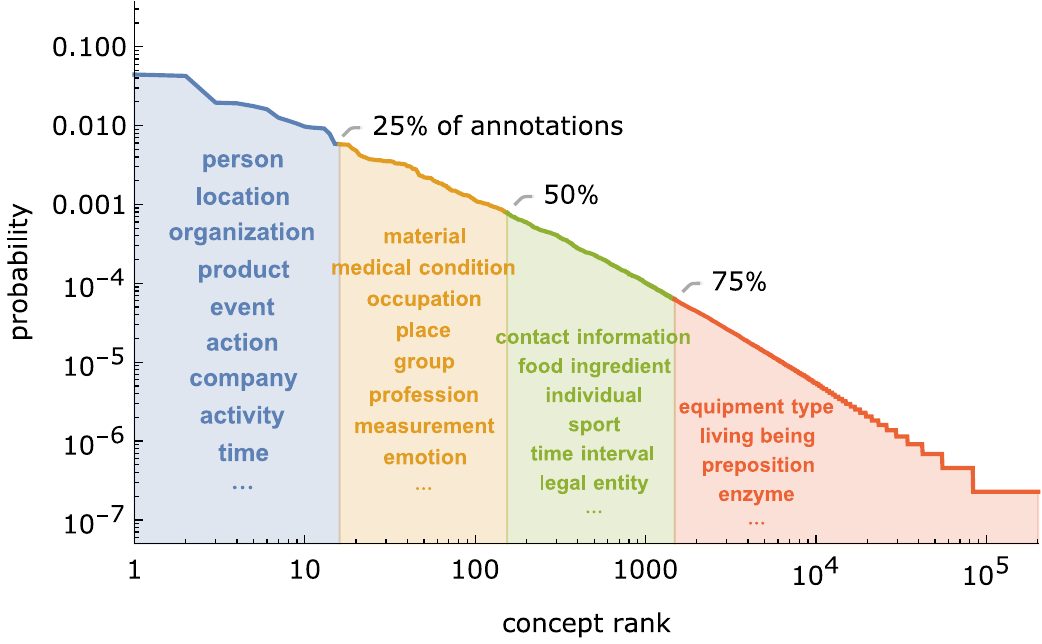}}
\vskip -0.1in
\caption{Frequency of each concept assigned by GPT-3.5, sorted from most to least common. We observe a heavy-tailed distribution.}
\label{distribution}
\end{center}
\vskip -0.1in
\end{figure}

We observe a high imbalance in concept frequencies. Common concepts such as ``person", ``location", or ``organization" each appears in more than 1\% of the extracted entities, while over 100k concepts are seen only once in the dataset. This heavy-tailed distribution of concept frequencies is shown in \cref{distribution}. We further investigate the importance of concept diversity in \cref{concept_diversity_section}.

\subsection{Model Training}
\label{Model_traninig}

We want to pre-train a language representation model using our annotated dataset, which contains 200k concepts and exhibits high concept-imbalance. Additionally, some concepts are similar to each other, such as ``company" and ``company name". Moreover, many potential entities in a sentence are not extracted. Due to these factors, training a conventional token classifier that only takes the text as input and returns a distribution over the entire set of concepts is not practical. We instead propose a training method based on the contrastive learning framework \citep{contrastive_learning}.

\begin{figure}[h]
\vskip 0.in
\begin{center}
\centerline{\includegraphics[width=\columnwidth]{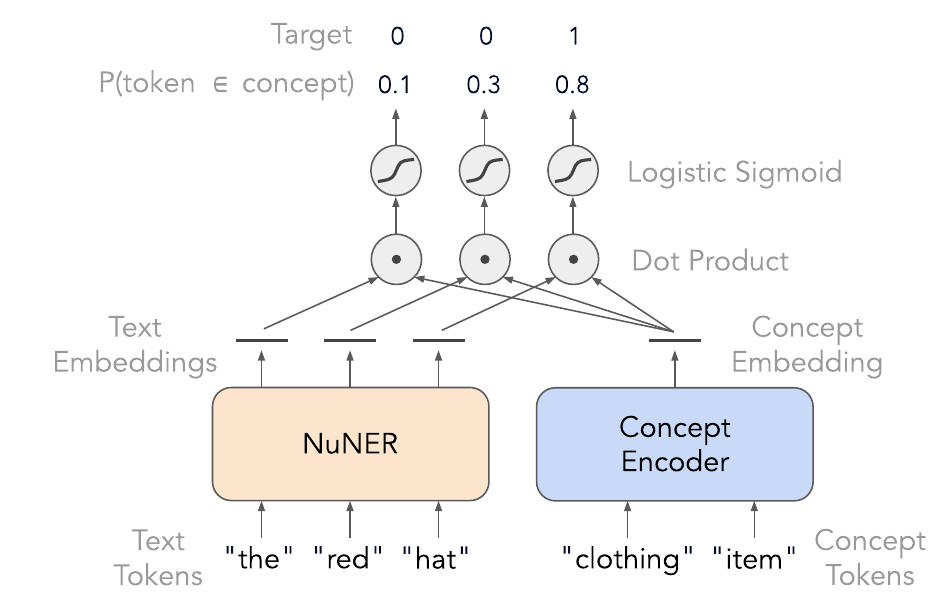}}
\caption{\NuNER{}'s pre-tranining procedure. The text and concept encoder are separated. Their embeddings are compared to obtain probabilities.}
\label{model-schema}
\end{center}
\vskip -0.1in
\end{figure}

\begin{figure*}
\vskip -0.1in
\begin{center}
\centerline{\includegraphics[width=.94\linewidth]{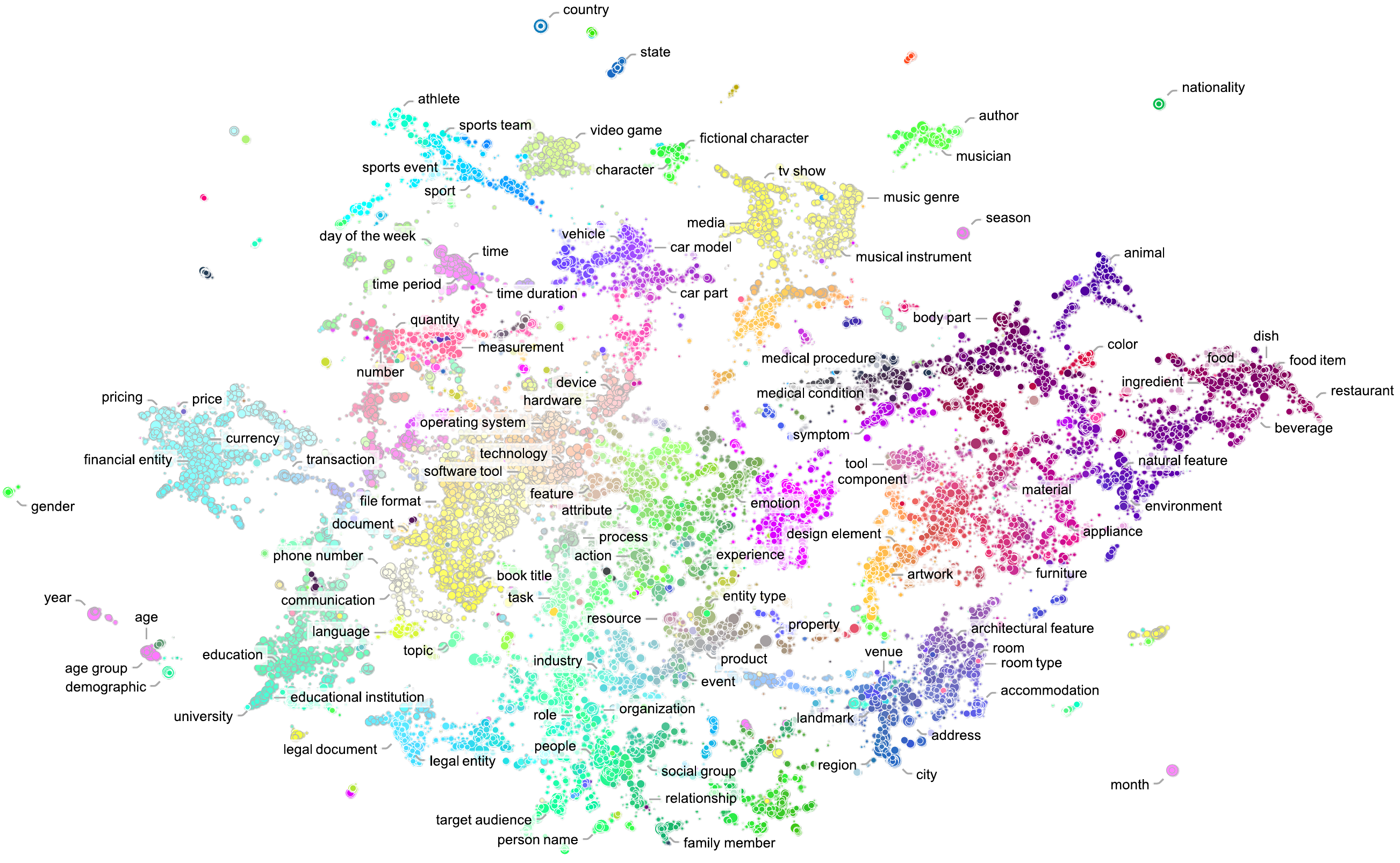}}
\caption{Feature map of the 50k most common concepts extracted by gpt-3.5-turbo-0301. Embeddings are obtained from the concept encoder (as depicted in \cref{model-schema}). UMAP\citep{mcinnes2020umap} is used to obtain 2D positions and 3D RGB color values. Disk size is proportional to log(concept frequency).}
\label{concept-map}
\end{center}
\vskip -0.2in
\end{figure*}

Our training network, depicted in \cref{model-schema}, consists of two separate sub-networks: The first is \NuNER{} --- the network of interest --- which encodes the input text as a sequence of vectors. The second encodes a concept name as a unique vector. The text vectors are matrix-multiplied with the concept vector to obtain logits, which are then passed through a logistic sigmoid to yield probabilities. During training, this setup encourages each token embedding to align with the concept embedding if the token instantiates the concept, and to become opposite otherwise.

For each training batch, we collect all the concepts exemplified in the batch and construct a binary array of dimensions \#sentences × \#tokens × \#concepts\_in\_batch. This array indicates the presence (1) or absence (0) of each concept present in the batch for every token in each sentence. We then use the binary cross-entropy loss to fit the network's probabilities to these target arrays.

Since we do not account for concepts not present in the batch, and because our dataset misses some concepts in some sentences, the probabilities generated by such a training network would need calibration for use in a zero-shot setting. However, in our case, this miscalibration is not a significant issue as we are only interested in using the text encoder \NuNER{}.

\subsection{Training Details}
We use RoBERTa-base \citep{roberta} for both the text encoder and the concept encoder. Our model is trained for 10 epochs on the full 1M sentence dataset where $90\%$ of sentences are used for training and $10\%$ for validation. We choose a learning rate $lr=0.00003$, batch size$ = 48$, and temperature before the sigmoid $\tau = 5$. We use AdamW optimizer \citep{adam} with $\beta1 = 0.9, \beta2 = 0.999, \epsilon=10^{-6}$, weight decay=$0.01$, and a linear scheduler with a warm-up for the first $10\%$ of the training steps. The bottom 6 layers of the text encoder are frozen as we found it leads to better training stability. After training, we discard the concept encoder and keep the text encoder \NuNER{}.

\section{Transfer Learning Performance}
\label{TLPerf}

\NuNER{} is designed to be fine-tuned to downstream NER problems in a data-efficient way. We are mostly interested in the transfer learning performance of \NuNER{} in an extended few-shot regime, typically from 1 to 100 annotations per entity types. We focus on this range because we believe it represents the level of annotation effort practitioners are willing to undertake without resorting to external annotation solutions, such as crowdsourcing.

\subsection{Few-Shot with Frozen Foundation}
\label{FSFrozen}

This first experiment aims to demonstrate the benefits of further pre-training an MLM encoder on our LLM-annotated dataset. Also, we want to compare \NuNER{}'s pre-training with an alternative large-scale NER dataset. To this end, we compare \NuNER{} with its base model, RoBERTa-base, as well as RoBERTa-base pre-trained on the dataset of NER-BERT \citep{ner-bert}, see \cref{related}.

We pre-train RoBERTa on NER-BERT data by replicating the training process of \citet{ner-bert}, with the exception that we freeze the bottom half of the network --- as when training \NuNER{} --- since we find it improves the final performance.

In order to compare these three foundation models, we transform them into token classifiers by attaching a linear layer on top of their final token representations. The entity types returned by the classifiers are mutually exclusive, and a special ``None" class is used to indicate the absence of entities. To simplify the few-shot training procedure --- and because our focus is on the relative performance of the models --- we train only the top layer while keeping the representation network frozen.

We use four datasets from different domains: OntoNotes 5.0 \citep{ontonotes}, BioNLP 2004 \citep{bionlp}, MIT Restaurant \citep{mit_rest}, and MIT Movie \citep{mit_rest}. Performance is measured using the macro-averaged F1-Score of token classifications.

We adopt the $k \sim 2k$ mining procedure of \citep{fewnerd} to obtain training examples that contain between $k$ and $2k$ annotations per entity type. We measure performance averaged over 10 training sets for each value of $k \in \{1, 2, 4, 8, 16, 32, 64\}$. The reported performance for a given $k$ is the average across all four datasets.

\begin{figure}
    \centering
     \vskip -0.in
    \begin{subfigure}{}
        \vskip -0.4in
        \centering
        \includegraphics[width=\columnwidth]{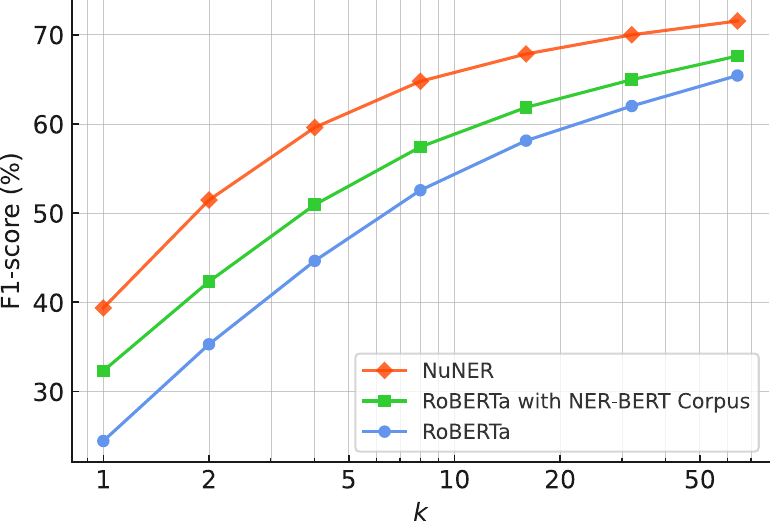}
       % \caption*{This is the figure} % Optional individual caption
    \end{subfigure}%
    \begin{subfigure}{}
        \vskip -0.15in
        \centering
        \begin{small}
        \begin{sc}
        \begin{tabular} {ccccc} % {lcccr}
        \toprule
        model & 1 & 4 & 16 & 64\\
        \midrule
        RoBERTa & 24.5 & 44.7 & 58.1 & 65.4 \\
        RoBERTa w. NER-BERT & 32.3 & 50.9 & 61.9 & 67.6 \\
        \NuNER{} & \textbf{39.4} & \textbf{59.6} & \textbf{67.8} & \textbf{71.5} \\
        \bottomrule
        \end{tabular}
        \end{sc}
        \end{small}
    %    \caption*{This is the table} % Optional individual caption
    \end{subfigure}
    \caption{Transfer learning performance of NuNER, RoBERTa, and RoBERTa pre-trained on NER-BERT data as function of $k$. \NuNER{} substantially outperforms both models for all training sizes. Full table and dataset-wise results are shown in \cref{full-f1-frozen-backbone} and \cref{NuNERvsRoBERTaALL} in the appendix.}
    \label{NuNERvsRoBERTa}
     \vskip -0.1in
\end{figure}

Performance is reported in \cref{NuNERvsRoBERTa}. As expected, \NuNER{} largely outperforms pure RoBERTa. More surprisingly, \NuNER{} outperforms RoBERTa trained on NER-BERT data by a large margin on all training sizes. We see this behavior in all four datasets, although the effect is stronger in some than others. This result demonstrates the benefits of pre-training using \NuNER{}'s dataset. 

\subsection{Few-Shot with TadNER on Few-NERD}
\label{few_nerd_sec}

\begin{table*}[h]
    \caption{Few-NERD performance using TadNER \citep{tadner} and a modified TadNER using \NuNER{}-BERT as the backbone.}
    \label{fewnerd-All}
    \vskip 0.15in
    \begin{center}
    \begin{small}
    \begin{sc}
    \begin{tabular}{cccccc}
    \toprule
    model & 5-way $1\sim2$  & 10-way $1\sim2$ & 5-way $5\sim10$  & 10-way $5\sim10$ &  avg \\
    \midrule
    TadNER | INTRA & 60.78±0.32 & 55.44±0.08 & 67.94±0.17 & 60.87±0.22 & 61.26 \\
    \NuNER{}-BERT | INTRA & \textbf{62.48±0.28} & \textbf{57.63±0.38} & \textbf{69.16±0.28} & \textbf{62.99±0.27} & \textbf{63.07} \\
    \midrule
    TadNER | INTER      & 64.83±0.14 & 64.06±0.19 & 72.12±0.12 & 69.94±0.15 & 67.74 \\
    \NuNER{}-BERT | INTER & \textbf{67.37±0.31} & \textbf{66.54±0.40}  & \textbf{73.50±0.09}  & \textbf{71.04±0.14} & \textbf{69.61} \\
    \bottomrule
    \end{tabular}
    \end{sc}
    \end{small}
    \end{center}
    \vskip -0.15in
\end{table*}

To complement the previous analysis, we evaluate \NuNER{} on Few-NERD \citep{fewnerd}, a challenging and widely recognized benchmark for few-shot NER. Our goal is to see whether \NuNER{} can achieve new state-of-the-art performance.

Few-NERD is comprised of 188k Wikipedia sentences that are human-annotated from a set of 8 coarse-grained and 66 fine-grained entity types. In this benchmark, the tested solution is allowed to undergo pre-training on a large subset of Few-NERD before being evaluated using 5,000 few-shot train-test splits. In the INTRA setting, the entities seen during pre-training and evaluation belong to different coarse-grained types, while in the INTER setting, the entities share the same coarse-grained types.

The current state-of-the-art on Few-NERD is TadNER \citep{tadner}, an advanced framework that employs a span-detection network, a type-classification network, as well as type-aware span filtering process and prototype construction. BERT is used for both networks. We adapt TadNER by replacing BERT with a \NuNER{} based on the same BERT. Furthermore, since \NuNER{} is designed to have all its entity-related knowledge in its last layer, we modify TadNER to only use this last layer instead of averaging over the last four layers.

Results are presented in \cref{fewnerd-All}. We observe that \NuNER{} outperforms the original TadNER results in all settings and training sizes, thereby establishing it as the new state-of-the-art for this benchmark. This further hints at the benefits of using \NuNER{}'s pre-training procedure.

\section{Ablation Studies}
\label{study}

We aim to understand which factors in the pre-training procedure most significantly affect the performance of \NuNER{}. To this end, we investigate the impact of text diversity, concept diversity, pre-training dataset size, and model size, using the same benchmark as in \cref{FSFrozen}.

\subsection{Effect of Text Diversity}
\label{diversity}

In \cref{FSFrozen}, we saw that \NuNER{}'s pre-training data, based on C4, leads to better performance than NER-BERT's pre-training data, based on Wikipedia. This might simply be because C4 is more diverse than Wikipedia.

To investigate this, we downsample both datasets to 50k sentences each, and annotate the Wikipedia subset with our LLM-annotation procedure. This way, the only distinguishing factor between these datasets is their original corpus: C4 vs. Wikipedia. We then pre-train RoBERTa on these two datasets and measure the transfer learning performance of the resulting models.

\begin{figure}
\vskip 0.in
\begin{center}
\centerline{\includegraphics[width=\columnwidth]{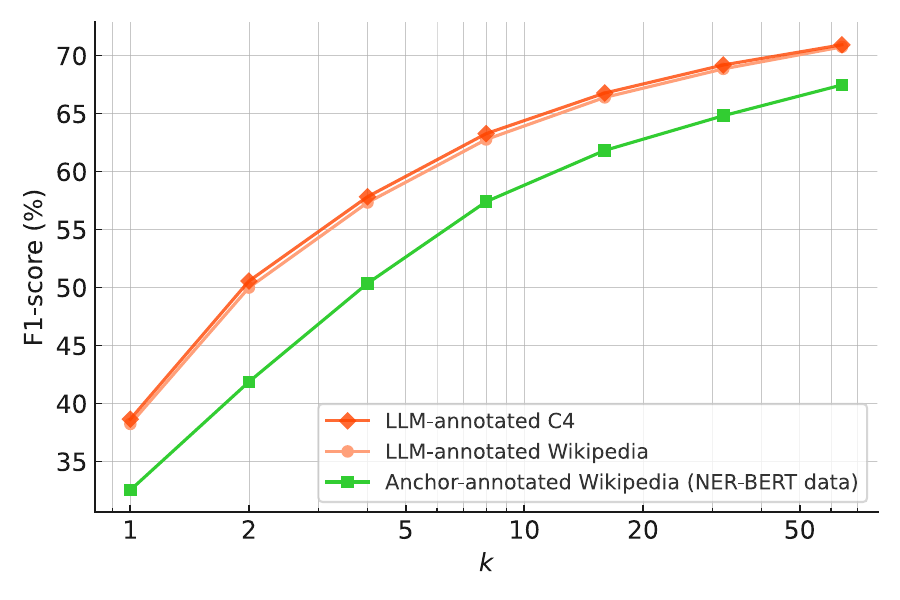}}
\vskip -0.15in
\caption{Effect of text diversity on \NuNER{}'s performance. Wikipedia and C4 lead to similar performance when they are both annotated by the LLM. Results table is shown in \cref{text-div-full} in the appendix.}
\label{TextDiversity}
\end{center}
\vskip -0.2in
\end{figure}

We see in \cref{TextDiversity} that the LLM-annotated C4 subset and the LLM-annotated Wikipedia subset result in very similar model performance. This shows that the main reason for the performance gap is the LLM-annotation procedure rather than the underlying corpus.

% This might be because the original RoBERTa has already been pre-trained on a highly diverse dataset \citep{roberta}, or because our benchmark is not out-of-domain enough with respect to Wikipedia.

\subsection{Effect of Concept Diversity}
\label{concept_diversity_section}

%We know from \cref{diversity} that the LLM-annotation procedure is the most important factor to explain \NuNER{}'s performance. We now want to know which aspect of this procedure helps most.

To understand the effect of concept diversity, we first take a random sample of 100k annotated examples from \NuNER{}'s dataset, which includes approximately 80k unique concepts. We then retain only the annotations from the top-$n$ most frequent concepts, simulating an annotation procedure that excludes rare concepts. We use $n=\text{4}$, 16, 154, 1.5k, and 80k concepts, corresponding to 12.5\%, 25\%, 50\%, 75\%, and 100\% of all annotations, respectively (see \cref{distribution}). We pre-train \NuNER{} on the resulting datasets and measure the transfer learning performance for $k=8$.

\begin{figure}[h]
\vskip 0.in
\begin{center}
\centerline{\includegraphics[width=\columnwidth]{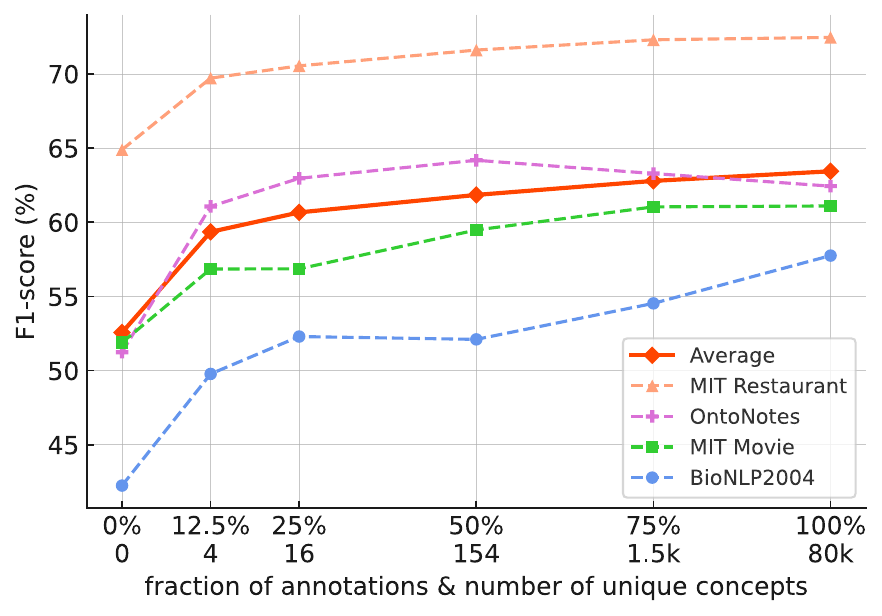}}
\vskip -0.1in
\caption{Effect of concept diversity on \NuNER{}'s performance. Results table is shown in \cref{conc-div-full} in the appendix.}
\label{ConceptDiversity}
\end{center}
\vskip -0.1in
\end{figure}

Results are shown in \cref{ConceptDiversity}. As expected, overall performance increases with concept diversity. However, there are variations across datasets. BioNLP appears to benefit the most from concept diversity while it seems to harm OntoNotes past 154 concepts. This difference is likely because BioNLP contains rarer concepts than OntoNotes. The performance degradation on OntoNotes may indicate the difficulty of encoding a large number of concepts into the 768-dimensional embedding vector.

Note that, in this experiment, the number of annotations grows with concept diversity, which might bring an additional effect. Results of \cref{datasizeSec} shows that such effect would account here for less than 1\% of F1-score.

%Interestingly, the largest gain in performance occurs between zero to four concepts. One reason for such gain is likely because these concepts are present in our benchmark dataset. It might also be due to the model learning how to detect entity spans with these types, which in turn helps extracting other entity types.

\subsection{Effect of Dataset Size}
\label{datasizeSec}

\begin{figure}[h]
\vskip 0.in
\begin{center}
\centerline{\includegraphics[width=\columnwidth]{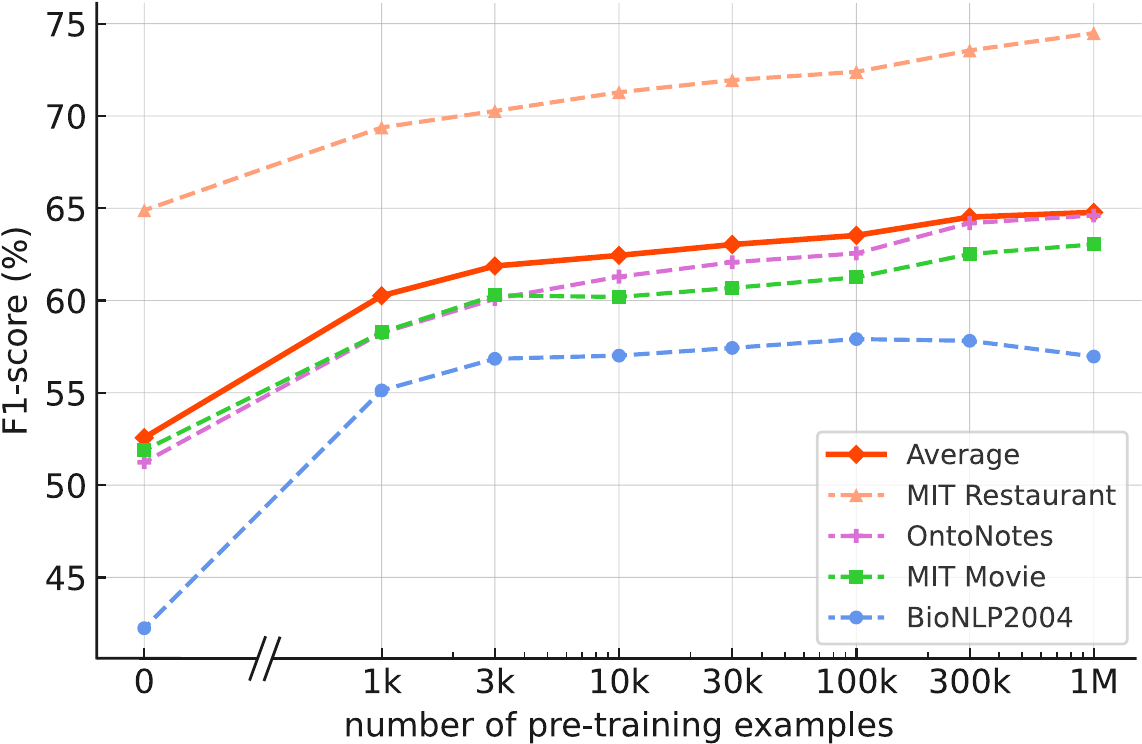}}
\vskip -0.1in
\caption{Effect of pre-training dataset size on \NuNER{}'s performance. Results table is shown in \cref{text-size-full} in the appendix.}
\label{datasize}
\end{center}
\vskip -0.2in
\end{figure}

We next investigate the effect of the pre-training dataset size, ranging from one 1k examples to 1M examples. We again pre-train \NuNER{} on each dataset and measure its transfer learning performance for $k=8$. Results are shown in \cref{datasize}.

As expected, the overall performance increases with data size, and continues to improve slightly from 300k to 1M examples. Again, we see variability across datasets, with the performance for BioNLP decreasing after 100k examples, while all other datasets experience a monotonic increase. The reason for this discrepancy is unclear.

% As in the concept-diversity study, we see the largest increase for the first 1k examples. This was unexpected to us and it would be interesting to understand why this happens with our data, but apparently not so much with NER-BERT data.

\subsection{Effect of Model Size}

\begin{figure}[h]
\vskip 0.in
\begin{center}
\centerline{\includegraphics[width=\columnwidth]{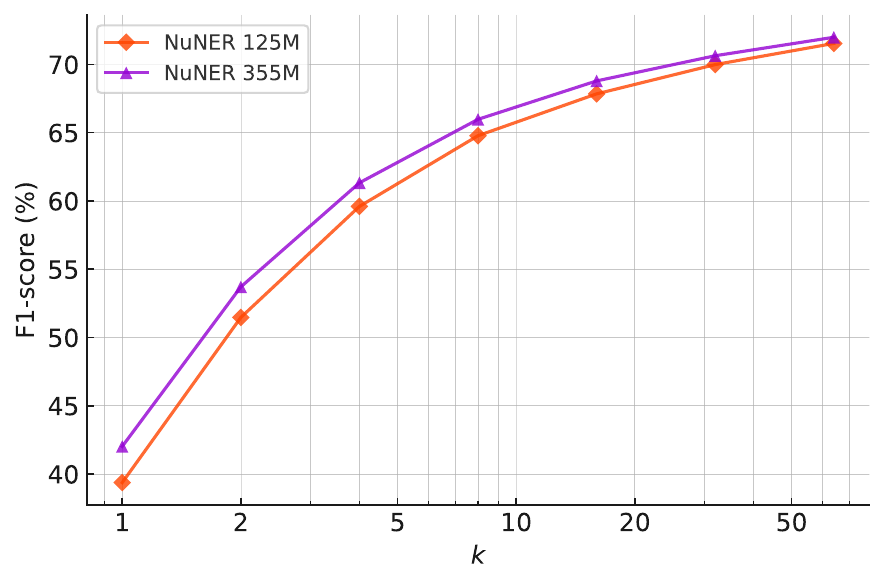}}
\vskip -0.15in
\caption{Effect of model size on \NuNER{}'s performance. Results table shown in \cref{full-model-size} in the appendix.}
\label{modelsize}
\end{center}
\vskip -0.2in
\end{figure}

Finally, we investigate the influence of model size. We pre-train a version of \NuNER{} using RoBERTa large (355M parameters) and compare it to the original \NuNER{} (155M parameters). We see in \cref{modelsize} an overall increase of the F1-score of a few percent. This increase is more pronounced for smaller training sets ($k < 10$) than for larger ones. Combined with the positive impacts of concept diversity and dataset size, this result suggests that scaling up both models and data would lead to further performance improvements.

\section{Comparison with LLMs}
\label{LLMSec}

Although the comparisons are indicative, we finally provide for information a comparison of \NuNER{} with modern generative LLMs. We choose GPT-3.5 (gpt-3.5-turbo-16k-0613) for its popularity and longer context, GPT-4 (gpt-4-0613) for its high performance \citep{zheng2023judging}, and UniversalNER (UniversalNER-7B-type) for its specialization in the NER task. UniversalNER has 56 times more parameters than \NuNER{}, and GPT-3.5 and GPT-4 are likely to have around 1,000 and 10,000 times more parameters than \NuNER{}, respectively.

GPT-4 and GPT-3.5 are used via in-context learning using Spacy's \href{https://github.com/explosion/spacy-llm/blob/09fcad1a6ebdb737e3daece08df19f54c1dcd531/spacy_llm/tasks/templates/ner.v3.jinja}{NER V3 prompt}. This advanced prompt template allows to create a prompt for a particular NER problem by providing it the set of entity types and some training examples.

UniversalNER \citep{zhou2023universalner} is trained to be used for zero-shot inferences, conducted through a conversation in which one sequentially prompt the model to identify each entity type. To adapt this model for a few-shot setting, we need to fine-tune it. We use the original training settings of UniversalNER but modify them to enhance few-shot learning performance, as detailed in \cref{UniNERTraining}. For \NuNER{}, we simply attach a two-layer fully-connected network regularized via dropout, and fine-tune the entire network for 30 epochs.

\begin{figure}[h]
\vskip 0.in
\begin{center}
\centerline{\includegraphics[width=\columnwidth]{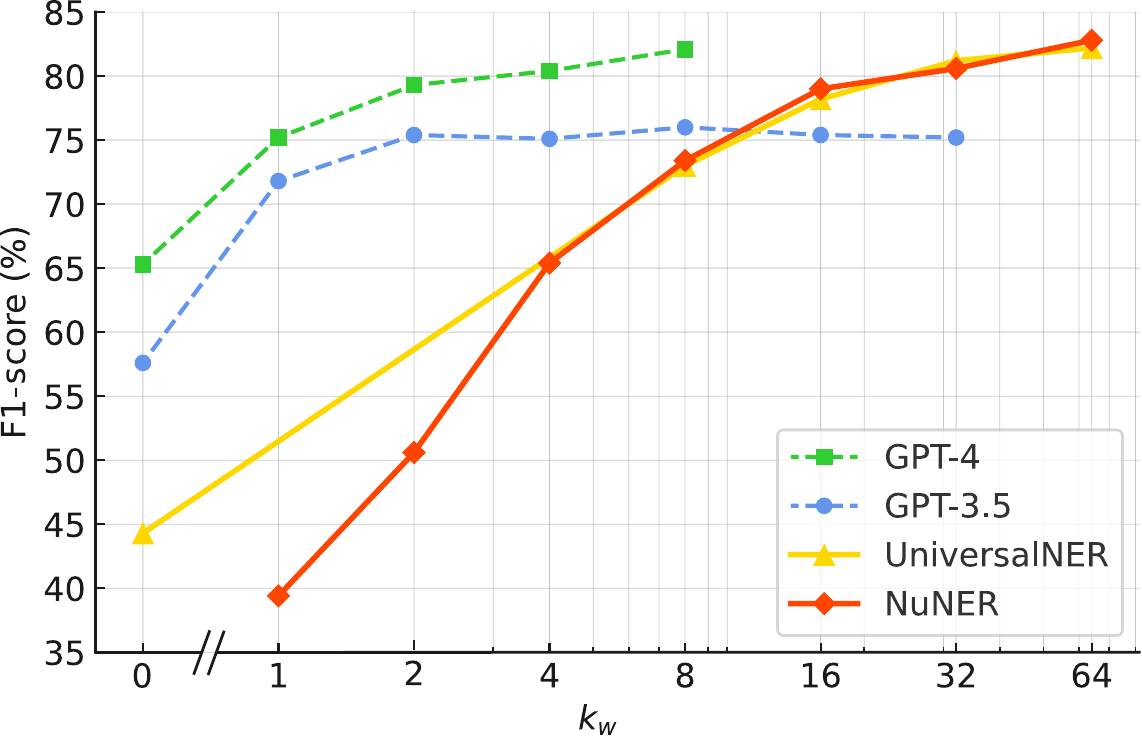}}
\vskip -0.05in
\caption{Comparison of \NuNER{} with LLMs. Dashed curves indicate in-context learning and solid curves indicate fine-tuning. Results tables are shown in \cref{mit_rest_macro} and \cref{bionlp_macro} in the appendix.}
\label{LLM}
\end{center}
\vskip -0.1in
\end{figure}

%NOTE: this part is a mess to explain. No way around it. %
Because of financial constraints associated with GPT-4, we deviate from the extended few-shot learning protocol of \cref{FSFrozen}. We only use the MIT Restaurant and BioNLP datasets, and downsample test sets to 1,000 examples. Also, we create training sets with a specific number of words belonging to a given entity type, that we call $k_w$, instead of using the $k \sim 2k$ entity-based mining method. We conduct several runs for each training-size (using the same training sets for all models) and average results, except for GPT-4 where we only perform one run. Results are presented in \cref{LLM}.

We find that GPT-3.5 and GPT-4 already perform well in the zero-shot regime, and show rapid improvement when examples are added to the prompt. However, for larger training sets, we see something unexpected: the performance of GPT-3.5 quickly plateaus, and it ends up being outperformed by both UniversalNER and \NuNER{} when $k_w>8$. The same thing might happen with GPT-4 but we cannot conclude with such noisy and incomplete results. UniversalNER starts lower than both GPT-3.5 and GPT-4, but steadily catches on to eventually surpass GPT-3.5. Clearly, fine-tuning does not suffer from this early saturation issue.

\NuNER{} begins at a lower performance level than the others as it is not intended to be a zero-shot model. However, it quickly matches UniversalNER's performance, and eventually surpass GPT-3.5. It remains unclear whether it would also surpass GPT-4.

To compare \NuNER{} and UniversalNER in a more standard and reproducible manner, we conduct an additional experiment using the $k \sim 2k$ setting on all four datasets mentioned in \cref{FSFrozen}. In this experiment, we use the full test sets and measure entity-level micro F1-score, averaged over these datasets. We perform 3 runs for each $k$. We see in \cref{NuNERUniNER} that \NuNER{} and UniversalNER have similar performance.

\begin{table}[h]
    \caption{\NuNER{} vs. UniversalNER few-shot entity-level F1-score in the $k \sim 2k$ setting showing similar performance. Dataset-wise tables can be found in \cref{F1-entity-dataset-8} and in \cref{F1-entity-dataset-64} in the appendix.}
    \label{NuNERUniNER}
    \vskip 0.15in
    \begin{center}
    \begin{small}
    \begin{sc}
    \begin{tabular} {ccccc} % {lcccr}
    \toprule
    model & $8 \sim 16$ & $64 \sim 128$ \\
    \midrule
    UniversalNER & $57.89 \pm 4.34$ & $71.02 \pm 1.53$ \\
    \NuNER{} & $58.75 \pm 0.93$ & $70.30 \pm 0.35$ \\
    \bottomrule
    \end{tabular}
    \end{sc}
    \end{small}
    \end{center}
    \vskip -0.in
\end{table}

The fact that \NuNER{} surpasses GPT-3.5 and possibly also GPT-4 is likely attributable to the limitations of in-context learning. The situation might differ with proper, albeit challenging, few-shot fine-tuning for these LLMs. More surprising to us is that \NuNER{} achieves performance comparable to UniversalNER despite being 56 times smaller and trained on similar data. This could be due to an inherent advantage of encoders over generative models for this task. Alternatively, it might be related to \NuNER{}'s pre-training procedure, which encourages human concepts to emerge in the last layers of the network, being easily accessible during few-shot training. Further experiments would be needed to explore this aspect.

\section{Conclusion}

Modern large language models are opening new possibilities for addressing traditional NLP tasks. We have introduced a procedure that uses these LLMs to create a compact, yet data-efficient, NER-specific foundation model. We foresee an increasing trend in the development of such task-specific foundation models, which will facilitate the creation of high-quality custom NLP models without requiring intensive human or computational resources.

\newpage
\newpage

\section*{Impact Statement}

This paper presents work whose goal is to advance the field of Machine Learning. There are many potential societal consequences of our work, none which we feel must be specifically highlighted here.


\bibliography{example_paper}
\bibliographystyle{icml2024}

%%%%%%%%%%%%%%%%%%%%%%%%%%%%%%%%%%%%%%%%%%%%%%%%%%%%%%%%%%%%%%%%%%%%%%%%%%%%%%%
%%%%%%%%%%%%%%%%%%%%%%%%%%%%%%%%%%%%%%%%%%%%%%%%%%%%%%%%%%%%%%%%%%%%%%%%%%%%%%%
% APPENDIX
%%%%%%%%%%%%%%%%%%%%%%%%%%%%%%%%%%%%%%%%%%%%%%%%%%%%%%%%%%%%%%%%%%%%%%%%%%%%%%%
%%%%%%%%%%%%%%%%%%%%%%%%%%%%%%%%%%%%%%%%%%%%%%%%%%%%%%%%%%%%%%%%%%%%%%%%%%%%%%%

\newpage
\newpage

\appendix
%\onecolumn
\section{Appendix}

\subsection{Comparison with LLMs - Details}

In the experiment of \cref{LLM} we only conduct one training run for GPT-4 for cost reasons. For all the other models we conduct 16 training runs for $k_w=1$, 8 training runs for $k_w=2$, 4 training runs for $k_w=4$, 2 training runs for $k_w=8$, and a unique training run for higher training sizes. These choices were made to reduce GPT-3.5 costs and to for the training data to be kept consistent across all models for each training size.

\subsection{UniversalNER Training Details}
\label{UniNERTraining}

We create UniversalNER's training data following \citet{zhou2023universalner}'s procedure: We use the dataset-specific instruction tuning template and populate it to extract all entity types. We don't need to add any negative sampling here since all possible entity types are queried, therefore having an empty list when no entities of a given type are present in the example.

We fine-tune UniversalNER using authors' fine-tuning code, only changing the number of training epochs and the number of gradient accumulation steps to obtain better few-shot results. After experimenting with external datasets, we found that using 20 epochs for $k_w=8$ and $k=8$, 15 epochs for $k_w=16$ and $k_w=32$, and 10 epoch for $k_w=64$ and $k=64$ worked well. We used these values and, for $k_w=8$ and $k=8$, also reduced the number of gradient accumulation steps to 4. We didn't train UniversalNER on $k_w<8$ as regularization was difficult to tune.

We also tried adopting LORA \citep{hu2021lora} for fine-tuning UniversalNER. Although, this methodology was more stable thanks to its implied regularization, it consistently led to worse results than full fine-tuning and required more time to converge.

We believe our heuristics allow to train UniversalNER pretty well in a few-shot setting, and are enough to make a rough comparison with \NuNER{}. However, we should note that fine-tuning such large model on such small amount of data is not easy, and there are certainly better automatic machine learning procedures for this model.

\subsection{Extra Tables and Figures}

In this section we present the tables that were partially cut or just illustrated as plots in the body of the paper. We also show dataset-wise results that were shown as averages in the body of the paper.

\begin{figure*}[ht]
\vskip 0.in
\begin{center}
\centerline{\includegraphics[width=0.9\textwidth]{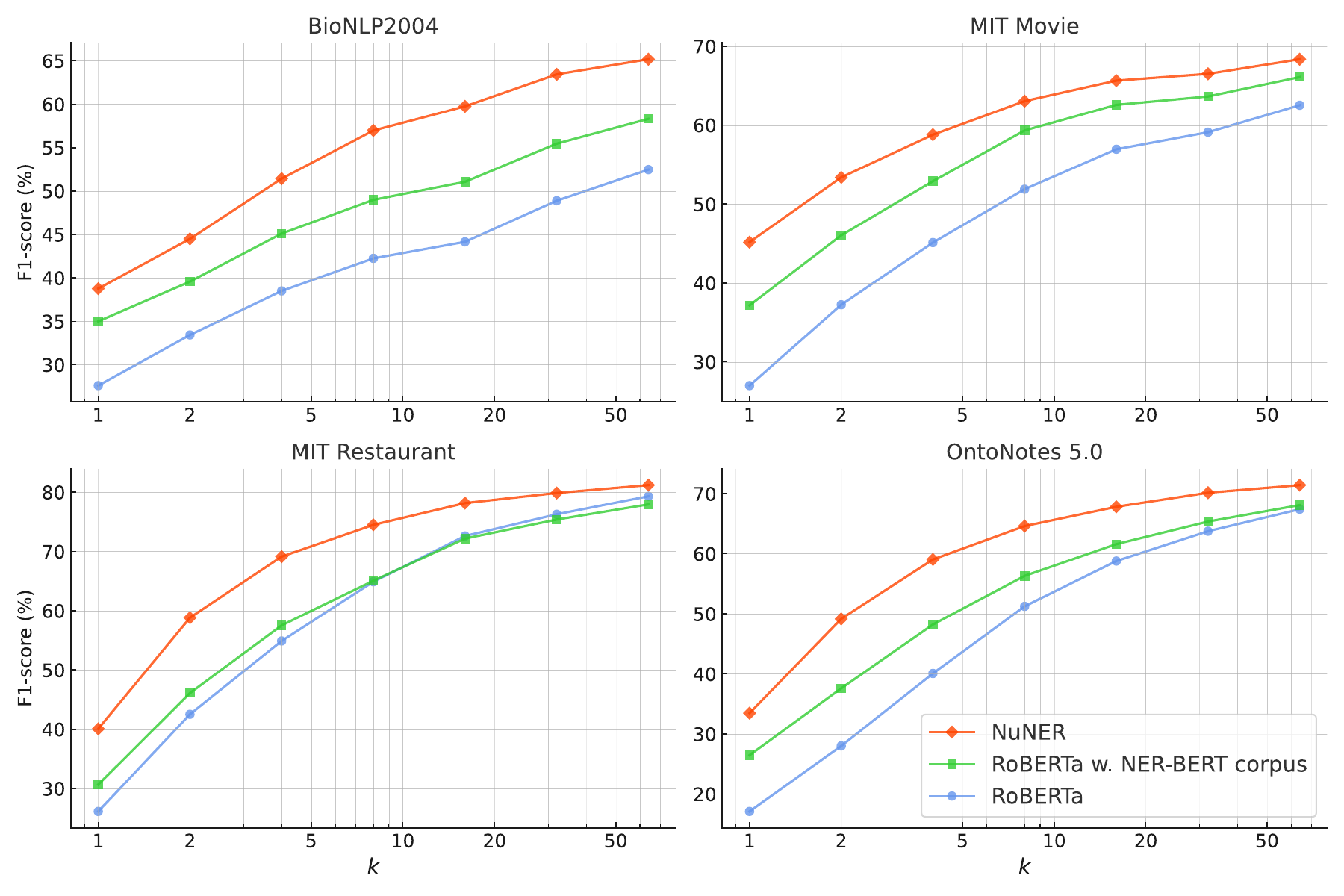}}
\vskip -0.15in
\caption{Dataset-wise results of \cref{NuNERvsRoBERTa}. Transfer learning performance of \NuNER{}, RoBERTa, and RoBERTa pre-trained on NER-BERT data as function of $k$. \NuNER{} substantially outperforms both models for all training sizes. We see similar behaviors for every dataset.}
\label{NuNERvsRoBERTaALL}
\end{center}
\vskip -0.1in
\end{figure*}

\begin{table}[t]
    \caption{Results table of \cref{NuNERvsRoBERTa}. Transfer learning performance of \NuNER{}, RoBERTa, and RoBERTa pre-trained on NER-BERT data as function of $k$. \NuNER{} substantially outperforms both models for all training sizes.}
    \label{full-f1-frozen-backbone}
    \vskip 0.15in
    \begin{center}
    \begin{small}
    \begin{sc}
    \begin{tabular}{ccccc}
    \toprule
    $k$ & RoBERTa & RoBERTa w. NER-BERT & NuNER \\
    \midrule
    1 & 24.46 & 32.32 & \textbf{39.37} \\ 
    2 & 35.30 & 42.33 & \textbf{51.46} \\ 
    4 & 44.65 & 50.94 & \textbf{59.60} \\ 
    8 & 52.56 & 57.42 & \textbf{64.78} \\ 
    16 & 58.12 & 61.85 & \textbf{67.84} \\ 
    32 & 62.00 & 64.96 & \textbf{69.98} \\ 
    64 & 65.42 & 67.61 & \textbf{71.53} \\ 
    \bottomrule
    \end{tabular}
    \end{sc}
    \end{small}
    \end{center}
    \vskip -0.1in
\end{table}

\begin{table}[t]
    \caption{Results table of \cref{TextDiversity}. Effect of text diversity on \NuNER{}'s performance. Wikipedia and C4 lead to similar performance when they are both annotated by the LLM.}
    \label{text-div-full}
    \vskip 0.15in
    \begin{center}
    \begin{small}
    \begin{sc}
    \begin{tabular}{ccccc}
    \toprule
    $k$ & Anch-An. Wiki & LLM-an. Wiki & LLM-an. C4 \\
    \midrule
    1 & 32.52 & 38.23 & \textbf{38.64} \\
    2 & 41.86 & 49.99 & \textbf{50.55} \\
    4 & 50.35 & 57.32 & \textbf{57.82} \\
    8 & 57.40 & 62.76 & \textbf{63.26} \\
    16 & 61.82 & 66.39 & \textbf{66.76} \\
    32 & 64.81 & 68.85 & \textbf{69.18} \\
    64 & 67.46 & 70.74 & \textbf{70.93} \\
    \bottomrule
    \end{tabular}
    \end{sc}
    \end{small}
    \end{center}
    \vskip -0.1in
\end{table}

\begin{table}[t]
    \caption{Results table of \cref{ConceptDiversity}. Effect of concept diversity on \NuNER{}'s performance.}
    \label{conc-div-full}
    \vskip 0.15in
    \begin{center}
    \begin{small}
    \begin{sc}
    \begin{tabular}{cccccc}
    \toprule
    $k$ & 4 & 16 & 154 & 1500 & 80k \\
     & 12.5\% & 25\% & 50\% & 75\% & 100\% \\
    \midrule
    1 & 32.78 & 36.95 & \textbf{39.56} & 39.45 & 38.83 \\
    2 & 44.40 & 47.33 & 49.75 & 50.53 & \textbf{50.69} \\
    4 & 52.57 & 54.75 & 56.60 & 57.58 & \textbf{57.97} \\
    8 & 59.35 & 60.66 & 61.84 & 62.78 & \textbf{63.43} \\
    16 & 63.74 & 65.05 & 65.49 & 66.38 & \textbf{66.83} \\
    32 & 66.76 & 67.67 & 67.96 & 68.92 & \textbf{69.27} \\
    64 & 69.31 & 70.02 & 70.03 & 70.76 & \textbf{71.04} \\
    \bottomrule
    \end{tabular}
    \end{sc}
    \end{small}
    \end{center}
    \vskip -0.1in
\end{table}

\begin{table}[t]
    \caption{Full results table of \cref{datasize} for all training sizes. Effect of pre-training dataset size on \NuNER{}'s performance.}
    \label{text-size-full}
    \vskip 0.15in
    \begin{center}
    \begin{small}
    \begin{sc}
    \begin{tabular}{cccccccc}
    \toprule
    $k$ & 1K & 3K & 10K & 30K & 100K & 300K & 1M \\
    \midrule
    1 & 32.8 & 35.6 & 37.3 & 38.3 & 38.9 & 39.3 & \textbf{39.4} \\
    2 & 45.5 & 48.0 & 49.3 & 50.2 & 50.8 & \textbf{51.6} & 51.5 \\
    4 & 54.0 & 56.2 & 56.9 & 57.6 & 58.1 & 59.1 & \textbf{59.6} \\
    8 & 60.3 & 61.9 & 62.5 & 63.0 & 63.5 & 64.5 & \textbf{64.8} \\
    16 & 64.3 & 65.4 & 66.1 & 66.6 & 66.9 & 67.7 & \textbf{67.8} \\
    32 & 66.5 & 67.8 & 68.4 & 69.0 & 69.3 & 69.8 & \textbf{70.0} \\
    64 & 68.3 & 69.4 & 70.0 & 70.7 & 71.1 & \textbf{71.6} & 71.5 \\
    \bottomrule
    \end{tabular}
    \end{sc}
    \end{small}
    \end{center}
    \vskip -0.1in
\end{table}

\begin{table}[t]
    \caption{Full results table of \cref{modelsize} for all training sizes. Effect of model size on \NuNER{}'s performance.}
    \label{full-model-size}
    \vskip 0.15in
    \begin{center}
    \begin{small}
    \begin{sc}
    \begin{tabular}{ccc}
    \toprule
    $k$ & NuNER & NuNER-large \\
    \midrule
    1 & 39.37 & 42.02 \\ 
    2 & 51.46 & 53.70 \\ 
    4 & 59.60 & 61.33 \\ 
    8 & 64.78 & 65.97 \\ 
    16 & 67.84 & 68.79 \\ 
    32 & 69.98 & 70.63 \\ 
    64 & 71.53 & 71.99 \\ 
    \bottomrule
    \end{tabular}
    \end{sc}
    \end{small}
    \end{center}
    \vskip -0.1in
\end{table}

\begin{table}[t]
    \caption{Dataset-wise results of \cref{NuNERUniNER} for $k=8$.}
    \label{F1-entity-dataset-8}
    \vskip 0.15in
    \begin{center}
    \begin{small}
    \begin{sc}
    \begin{tabular}{ccccc}
    \toprule
    dataset & NuNER & UniversalNER \\
    \midrule
    bionlp & $43.84 \pm 2.18$ & $41.79 \pm 17.08$ \\
    mit movie & $58.69 \pm 1.16$ & $61.57 \pm 2.26$ \\
    mit restaurant & $62.66 \pm 2.28$ & $63.4 \pm 1.69$ \\
    ontonotes & $69.82 \pm 1.64$ & $64.8 \pm 1.17$ \\
    average & $58.75 \pm 0.93$ & $57.89 \pm 4.34$\\
    \bottomrule
    \end{tabular}
    \end{sc}
    \end{small}
    \end{center}
    \vskip -0.1in
\end{table}

\begin{table}[t]
    \caption{Dataset-wise results of \cref{NuNERUniNER} for $k=64$}
    \label{F1-entity-dataset-64}
    \vskip 0.15in
    \begin{center}
    \begin{small}
    \begin{sc}
    \begin{tabular}{ccccc}
    \toprule
    dataset & NuNER & UniversalNER \\
    \midrule
    bionlp & $61.18 \pm 1.15$ & $59.94 \pm 6.03$ \\
    mit movie & $65.56 \pm 0.29$ & $69.50 \pm 0.57$ \\
    mit restaurant & $73.66 \pm 0.65$ & $75.84 \pm 0.97$ \\
    ontonotes & $80.81 \pm 0.41$ & $78.78 \pm 0.26 $\\
    average & $70.30 \pm 0.35$ & $71.02 \pm 1.53$ \\
    \bottomrule
    \end{tabular}
    \end{sc}
    \end{small}
    \end{center}
    \vskip -0.1in
\end{table}

\begin{table}[t]
    \caption{Results table of \cref{LLM} for MIT Restaurant. Comparison of \NuNER{} with LLMs. F1-macro token classification metric.}
    \label{mit_rest_macro}
    \vskip 0.15in
    \begin{center}
    \begin{small}
    \begin{sc}
    \begin{tabular}{ccccc}
    \toprule
    $k_w$ & GPT-3.5 & GPT-4 & UniversalNER & NuNER \\
    \midrule
    0 & 51.70 & \textbf{69.54} & 28.35 & - \\
    1 & 77.26 & \textbf{79.66} & - & 38.48 \\
    2 & 78.83 & \textbf{81.48} & - & 51.08 \\
    4 & 80.01 & \textbf{84.54} & - & 74.47 \\
    8 & 81.00 & \textbf{85.35} & 76.56 & 78.67 \\
    16 & \textbf{82.00} & - & 80.30 & 81.58 \\
    32 & 81.34 & - & 82.36 & \textbf{82.53} \\
    64 & - & - & 83.40 & \textbf{84.79} \\
    \bottomrule
    \end{tabular}
    \end{sc}
    \end{small}
    \end{center}
    \vskip -0.1in
\end{table}

\begin{table}[t]
    \caption{Results table of \cref{LLM} for BioNLP. Comparison of \NuNER{} with LLMs. F1-macro token classification metric.}
    \label{bionlp_macro}
    \vskip 0.15in
    \begin{center}
    \begin{small}
    \begin{sc}
    \begin{tabular}{ccccc}
    \toprule
    $k_w$ & GPT-3.5 & GPT-4 & UniversalNER & NuNER \\
    \midrule
    0 & \textbf{63.48} & 61.09 & 60.23 & - \\
    1 & 67.62 & \textbf{70.82} & - & 40.39 \\
    2 & 70.70 & \textbf{77.10} & - & 50.02 \\
    4 & 70.22 & \textbf{76.32} & - & 56.31 \\
    8 & 70.86 & \textbf{78.80} & 69.52 & 68.01 \\
    16 & 68.88 & - & 76.09 & \textbf{76.31} \\
    32 & 69.12 & - & \textbf{80.09} & 78.78 \\
    64 & - & - & \textbf{80.96} & 80.85 \\

    \bottomrule
    \end{tabular}
    \end{sc}
    \end{small}
    \end{center}
    \vskip -0.1in
\end{table}

\end{document}